\documentclass[conference]{IEEEtran}

\IEEEoverridecommandlockouts
\usepackage{cite}
\usepackage{amsmath}
\usepackage{float}
\usepackage{caption} 

\usepackage{graphicx}
\usepackage{tabularx} 
\usepackage{booktabs} 
\usepackage{amsmath,amssymb,amsfonts}
\usepackage{algorithmic}
\usepackage{graphicx}
\usepackage{textcomp}
\usepackage{xcolor}
\def\BibTeX{{\rm B\kern-.05em{\sc i\kern-.025em b}\kern-.08em
    T\kern-.1667em\lower.7ex\hbox{E}\kern-.125emX}}

\begin{document}

\title{Fight Scene Detection for Movie Highlight Generation System }

\author{
\IEEEauthorblockN{Aryan Mathur}
\IEEEauthorblockA{Undergraduate at Indian Institute of Technology, Palakkad, India\\\ aryannmathur@gmail.com}
}
\maketitle

\begin{abstract}
In this paper of research based project, using Bidirectional Long Short-Term Memory (BiLSTM) networks, we provide a novel Fight Scene Detection (FSD) model which can be used for Movie Highlight Generation Systems (MHGS) based on deep learning and Neural Networks . Movies usually have Fight Scenes to keep the audience amazed.  For trailer generation, or any other application of Highlight generation, it is very tidious to first identify all such scenes manually and then compile them to generate a highlight serving the purpose. Our proposed FSD system utilises temporal characteristics of the movie scenes and thus is capable to automatically identify fight scenes. Thereby helping in the effective production of captivating movie highlights.

We observe that the proposed solution features 93.5\% accuracy and is higher than 2D CNN with Hough Forests which being 92\% accurate and is significantly higher than 3D CNN which features an accuracy of 65\%.

Key Terms: Fight Scene Detection, Violence Classification, Highlight Generation, Bidirectional Long Short Term Memory (Bi-LSTM), Deep Learning, Convolutional Neural Network
\end{abstract}

\section{Introduction}
The highlights this model generates from movies include all the violent scenes. Creating such highlights provides users with an idea of how much violent the movie is, gives an idea of the key scenes to look out for and can be used in generating AI based trailers.

\subsection{Significance of the Project}
The significance of  Fight Scene Detection (FSD) framework for Movie Highlight Generation revolves around the  way movie highlights are generated. This project's following major applications add to its' significance:

\begin{itemize}
    \item 
Gaining a perspective:  Movie highlights can help people decide whether to watch a particular movie or not, in particular whether to show it to their young kids or not by seeing the intensity and amount of violence in the film.  These highlights also give an idea about what to expect from the movie.

    \item 
Trailer generation:  This Movie highlight Generation Model can help filmmakers generate trailers more efficiently by reducing the time and efforts editors spend in extracting violence based scenes to use them in the trailers.The Fight Scene Detection (FSD) model optimises the  process of highlight generation concerning Fight Scenes in particular. Hence  reducing the time and manual effort required for extracting violent scenes from a movie.  It therefore offers scalability and cost effectiveness to the media houses.

\item Multidisciplinary servings: Combining the techniques of  deep learning , computer vision with the specific challenges of  content generation targeting violent scenes, this project serves to both academic research and industrial applications. Deep learning applications, such as Bidirectional Long Short-Term Memory (BiLSTM) networks,  can do  multimedia analysis tasks in the real world. This FSD model appears to be  a practical solution for creators, and media houses to improvise the highlight generation process and serve their content better to the target audience.
\item This project can also be used in security applications. For an instance , feeding in the live feed from CCTV cameras can give instantaneous alerts to police if violent situations are detected in any part of the city. Moreover this would give them a better idea of ongoing crime rate in the  town.
\item This project can also be used in Censor Boards. This would reduce their manual effort in going through movies and identifying intense violence scenes while filtering the movies.
\end{itemize}

\subsection{Survey of Existing methods and technologies}\label{AA}

The Movie Highlight Generation Systems (MHGS) works on the Fight Scene Detection (FSD) model created in this porject to  identify and extract key action sequences from movies. Upon research, the following already existing approaches were found:

\begin{enumerate}
    \item \textbf{Manual Annotation and Rule-based Approaches:} Done by humans, takes in a lot of effort , hardwork and time.
    \item \textbf{Feature-based Methods:} This method works on motion analysis along with audio analysis
    \item \textbf{Machine Learning Techniques:} Convolutional Neural Networks (CNNs) and Recurrent Neural Networks (RNNs) are used at certain places.
    \item \textbf{Temporal Modeling:} Temporal Convolutional Networks (TCNs) and Recurrent Neural Networks (RNNs) capture the depth of features in the data more effectively.
    \item \textbf{Hybrid Approaches:}audio and vidual features are studied together.
    
\end{enumerate}

\subsection{Problems and Statements}
Intensive human efforts in manual curation of movie highlights, including the identification of fight scenes and then   extraction , are both time consuming and exhaustive. It is also subjective to perspective of the person doing the task and is heavily prone to biases. Existing automated methods  are not scalable, are effort exhaustive and struggle with scene complexity. They also suffer from accuracy issues. There is a need for automated efficient and reliable Fight Scene Detection (FSD) models that can  identify and extract fight scenes from movies on their own to enhance the curation of Fight Scene  highlights in movies.

\subsection{Motivation}
In this AI driven era, when everything is getting automated, from the fans at your home to the recommendation algorithms when you open youtube, the motivation behind this research based project rises from the increasing demand for automated solutions in every sector to also enhance the process of creation of  highlights. Action films have fight scenes as their key moments. These hold significant attention of audiences and play a crucial role in deciding viewer engagement. Whereas, the manual curation of movie highlights, is labor-intensive, time-consuming,  subjective and often inaccurate.

With the development of Fight Scene Detection (FSD) algorithms, these challenges are addressed.  The process of movie highlight generation is also streamlines. Automated FSD improves efficiency by reducing the reliance on manual efforts. It also enhances the consistency, quality of the generated highlights. On a broader scale, it empowers  creators and streaming platforms to deliver personalized content to action loving based audiences worldwide.

\subsection{Major Objectives with Work Plan}
\subsection*{Major Objectives:}

\begin{enumerate}
    \item \textbf{Develop an Automated FSD Framework:} Design and implement a BiLSTM Model for Fight Scene Detection for  detecting, extracting and compiling fight scenes from movie sequences.
    
    \item \textbf{Improvised Scene Detection Accuracy:} Look for methods to improve the accuracy  of FSD algorithms on diverse cinematic contexts.
    
    \item \textbf{Evaluate  Performance of the model:} Conduct  evaluations to assess the effectiveness of the developed FSD model.
    \item \textbf{Functionality of Highlight Generation:} Making specific funtions on the model for this very purpose.

\end{enumerate}

\subsection*{Work Plan:}

\begin{enumerate}
    \item \textbf{Research work:} Conducted an extensive research on already available solutions and the applications of technologies that would prove useful in this project. i.e. Machine Learning, Deep Learning architectures, and multimedia analysis techniques.
    
    \item \textbf{Data Collection :} Gathered diverse movie datasets covering a range of genres and styles. Preprocess the data to extract relevant audiovisual features, segment movie sequences, and annotate fight scenes.
    Real-life Violence Situations Dataset available on Kaggle served this purpose.
    \item \textbf{Data Preprocessing :} The videos were labelled, frames were split, resized to 64x64 dimensions, from which features were extracted. 
    The paths to each class were defined and this new dataset was now saved in .npy format.
    For the entirity of this project, videos were processed at 16fps.

    \item \textbf{Model Architecture Development :} Designed and implemented the FSD architecture using deep learning architectures.
    
    \item \textbf{Performance Evaluation (2 months):} Evaluated the performance of the developed FSD model using  evaluation metrics on the testing dataset.
    
    \item \textbf{Optimization and Integration :} Optimized the performance of the model and used it in Movie Highlight Generation.

    \item  \textbf{GUI Development :} Developed a GUI using gradio.
    
    \item \textbf{Documentation and Dissemination :} Prepared a technical documentation, research papers summarizing the findings  of the research  project.
\end{enumerate}

\section{Materials, Methods and Modules}

\subsection{System Architecture with Description}

The Fight Scene detection Model's  architecture is a sequential set of following instructions to perform analysis on video data:

\begin{enumerate}
    \item Input Layer: The Input layer has the frames resized to the size called by \verb|IMAGE_HEIGHT| (64) and \verb|IMAGE_WIDTH| (64), and  3 color channels (RGB). It takes in sequences of 16 frames at a time. These 
    \item TimeDistributed MobileNet:Each frame of the input layer is processed using the \verb|TimeDistributed| wrapper to apply  MobileNet (FineTuning Model)  .This finetunes the model and captures spatial data.
    \item Dropout: Certain layers are added to remove a part of the input data that might be overfitting. A dropout of 0.3 was used to prevent overfitting.
    \item TimeDistributed Flatten: The data is then vectorised to a 1D vector while preserving its' features and temporal characteristics.
    \item Bidirectional LSTM: Bidirectional Long Short-Term Memory (LSTM) units capture temporal data of the video sequence in input. This feature lets the model learn from both past and future features of the dataset.
    \item Dense Layers with Dropout:ReLU activation is used on fully connected (Dense) layer for transformation of features. The Dropout layers are added once more after each Dense layer to not let overfitting happen.
    \item Output Layer: The soft-max activated on the final dense layer produces the  output, that predicts the class to which input data belongs from the CLASSES\_LIST.

\end{enumerate}

Therefore this model combines together CNN, RNN, BiLSTM, Dropout, ReLU, softmax to operate on the data as expected.
This model was trained on:
Total params: 3637090 (13.87 MB)
Trainable params: 3334818 (12.72 MB)
Non-trainable params: 302272 (1.15 MB)
making the last 60 layers trainable of the 4 GB dataset.

\subsection{Training}
This project utilised TensorFlow to train the model. With two callback functions of EarlyStopping and ReducedLRonPlateau.
\\
For early stopping with a patience factor of 10, the training began. Then it halts training when accuracy does not improve and restores the best weights. 
\\
The plateau callback then reduces patience factor to 5 and made the learning rate to a minimum of 0.00005. This function uses categoricalcrossentropy to deal with loss and stochastic gradient descent (sgd) optimizer\\
In all, the model is supposed to run 50 epochs on the training dataset. The model finally is trained on the training dataset with the validations as specified while using the two callback funtions.
\\
\\
The final epoch was epoch number 35 and gave the following evaluation parameters:
Epoch 35:\\
\\
Learning rate = 0.003600000031292438.\\
loss : 0.0283 \\accuracy: 0.9896 \\val\_loss: 0.1829 \\val\_accuracy: 0.9500 \\ lr: 0.0060

\begin{figure}
    \includegraphics[width=0.5\textwidth]{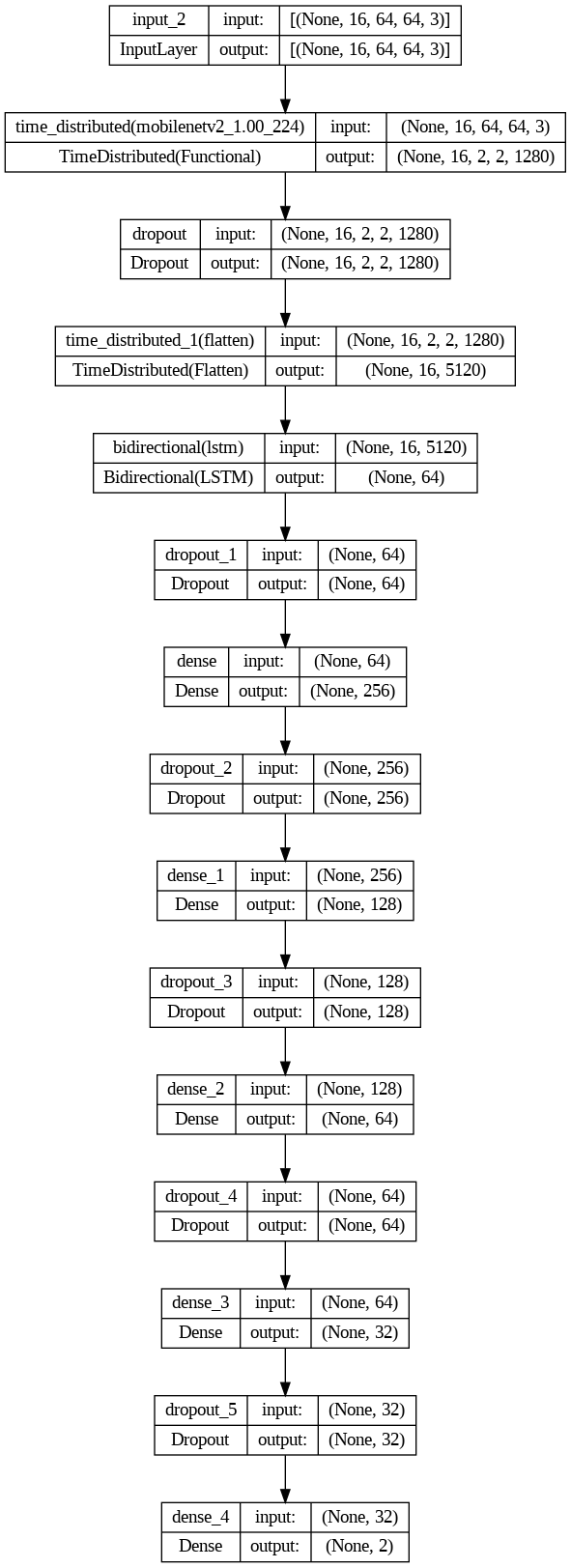}
    \caption{System Architecture}
    \label{Model Architeture}
\end{figure}

\subsection{System Specification Table}\label{AA}
The project was done on a system with the following specifications on a Google Colab environment as a Jupyter Notebook, utilising a cloud hosted GPU RAM of NVIDIA Tesla T4.
\begin{table}[H]
\centering
\begin{tabularx}{\textwidth}{lX}
\toprule
\textbf{Component} & \textbf{Specification} \\
\midrule
Processor & Apple M1 (8-core CPU, 8-core GPU) \\
Memory & 8 GB RAM (Unified Memory) \\
Storage & 512 GB SSD \\
Operating System & macOS \\
Colab GPU & NVIDIA Tesla T4 (shared, approximately 10 GB RAM) \\
\bottomrule
\end{tabularx}
\caption*{System Specifications}
\label{tab:system_specs}
\end{table}
\subsection{Description of Sensors or Other Modules Related to the Project}
Camera is the only sensor required for the project. We can use any HD Camera to capture a video and then transfer the video to a computer, feed it post processing to the model created in this project.
\subsection{Block Diagram or Flowchart with Description}

\begin{figure}[H]
    \includegraphics[width=0.5\textwidth]{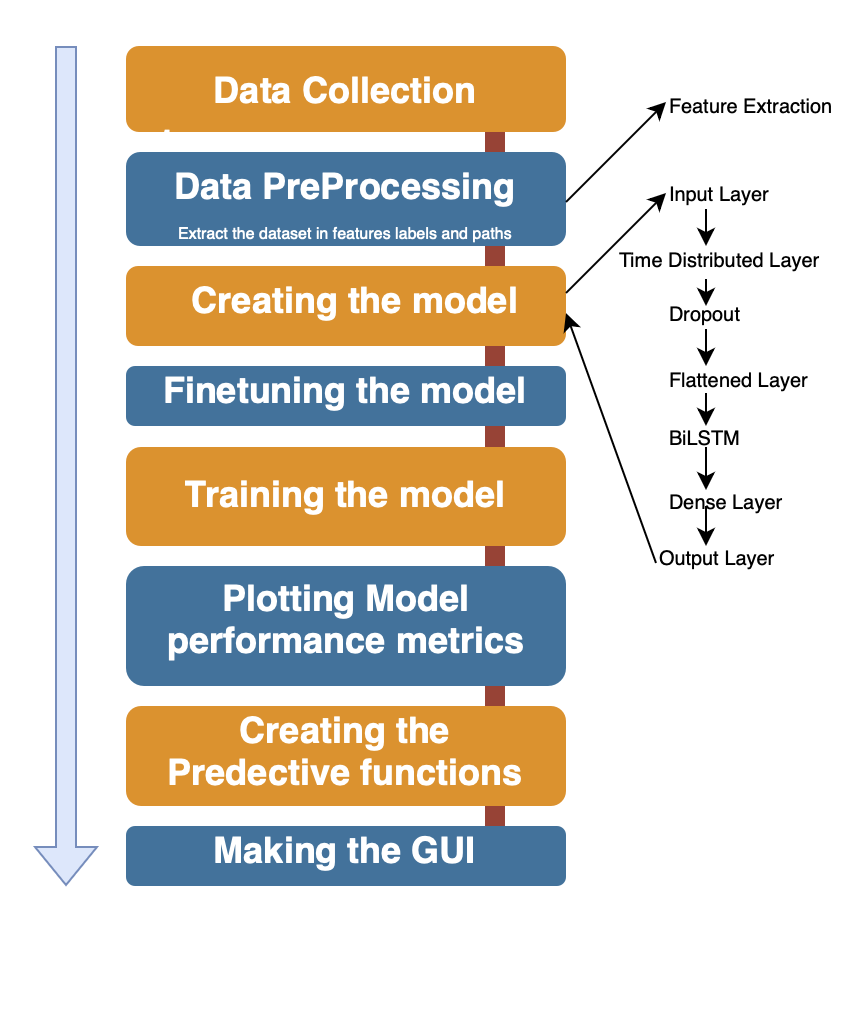}
    \caption{FlowChart}
    \label{Flowchart}
\end{figure}
As the flowchart suggests , the project started with data collection, which then had to be preprocessed to extract features, etc. Then the model was created with the given sequential instruction set. Then the model was finetuned with Mobilenet. We then plotted performance metrucs followed by predictive functions and then making the GUI using Gradio.
\subsection{}
\subsection{Mathematical Expressions Related to the Project Tasks}
Bidirectional Long Short-Term Memory (BiLSTM) is a  recurrent neural network (RNN) architecture which is used or sequence processing tasks.  Given a sequence of video frames $\{X_1, X_2, ..., X_T\}$, where $T$ is the number of frames in the sequence, the BiLSTM model predicts whether each frame corresponds to a fight scene or not.
Let $h_t^f$ and $h_t^b$ represent the forward and backward hidden states of the BiLSTM at time step $t$, respectively. The hidden states are computed as follows:
Forward pass:
\[
h_t^f = \text{LSTM}_f(X_t, h_{t-1}^f)
\]Backward pass:
\[
h_t^b = \text{LSTM}_b(X_t, h_{t+1}^b)
\]The final hidden state $h_t$ at each time step $t$ is a concatenation of the forward and backward hidden states:
\[
h_t = [h_t^f; h_t^b]
\]
To predict whether a frame is part of a fight scene, a softmax layer is applied to the output of the BiLSTM:
\[
\hat{y}_t = \text{softmax}(W_o h_t + b_o)
\]
where $\hat{y}_t$ is the predicted probability distribution over classes (fight scene or non-fight scene), $W_o$ and $b_o$ are the weights and bias of the output layer, respectively.

\subsection{User Interface Related to Project Tasks}\label{AA}
Having explored multiple options for UI generation including tkinter, flask, etc. Gradio came out to be the most user friendly and modern looking GUI creation module. Using gradio, the UI can also be hosted on a server and used as an API. The UI consists of an upload button that takes video from the user and a playback of the output (generated highlight of the fight scenes)
Developed GUI using Gradio. 
\begin{figure}[H]
    \includegraphics[width=0.5\textwidth]{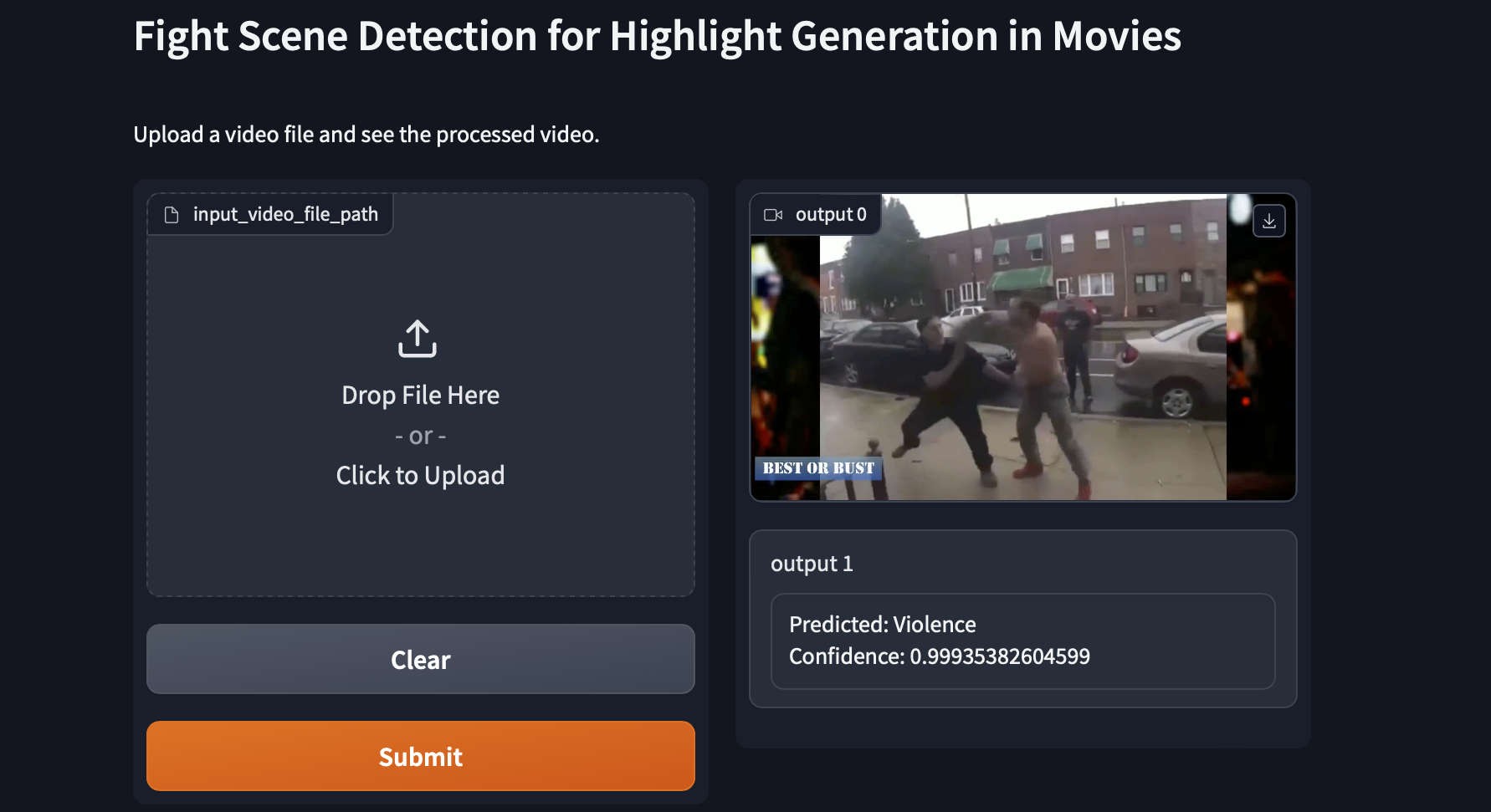}
    \caption{GUI- Dark Mode}
    \label{GUI Dark}
\end{figure}
\begin{figure}[H]
    \includegraphics[width=0.5\textwidth]{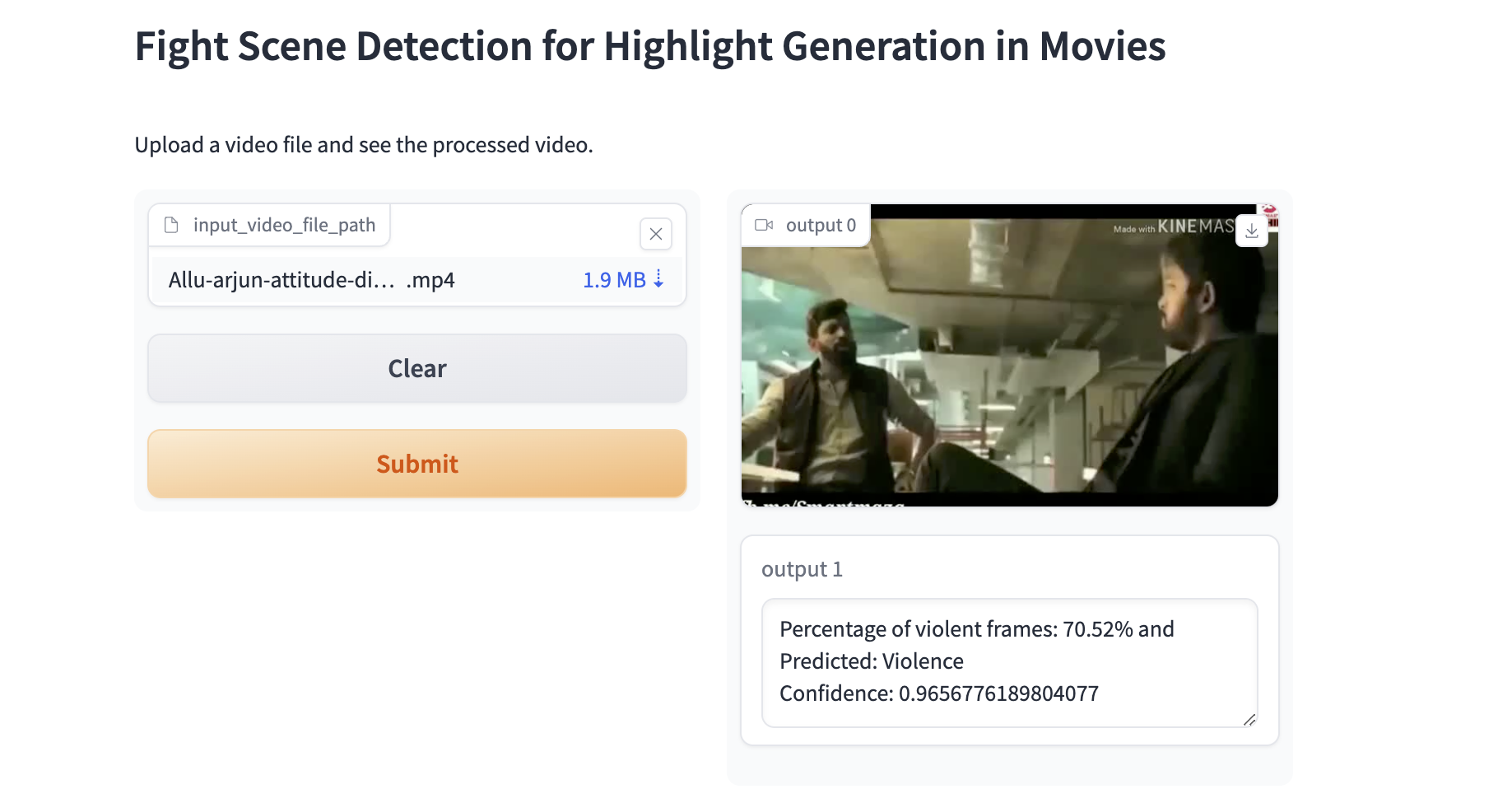}
    \caption{GUI- Light Mode}
    \label{GUI Light}
\end{figure}

\section{Results and Discussions}

\subsection{Performance Metrics}

The model works with an accuracy of 93.5 \% and a loss of 0.2266
We obtain the confusion matrix (Fig.4) of the modeling by testing the model o unseen dataset. The confusion matrix  has four metric parameters: True Positives (TP), True Negatives (TN), False Positives (FP), and False Negatives(FN). These are given below:
\begin{itemize}
\item
Sensitivity (SE) / Recall /: It measures the part of actual violent instances (violent events) that are correctly predicted by the model. It is given by
\[
SE = \frac{TP}{TP + FN}
\]
\item 
Specificity (SP) : It measures the fraction of actual non violent instances that are correctly predicted by the model. It is given by
\[
SP = \frac{TN}{TN + FP}
\]
\item
Overall Accuracy(ACC): It measures the proportion of correct predictions (which are violent and predicted violent and which are non violent and predicted nonviolent) against the total predictions made by the model. It is given by
\[
ACC = \frac{TN + TP}{TN + TP + FN + FP}
\]
\item 
Precision: It measures the part of  Violent instances that are actually violent and predicted violent. It is like an accuracy but only for the positive(Violent cases in this case).
\[
PE = \frac{TP}{TP + FP}
\]
\item 
F1 Score: It is a performance metric used for evaluating a model. It is calculated from the Precision and Sensitivity. It is given by
\[
F1 = 2 \times \frac{PE \times SE}{PE + SE}
\]

\end{itemize}
\begin{table}[H]
\centering
\begin{tabular}{lr}
\toprule
\textbf{Performance Metric (Out of 1)}& \textbf{Value} \\
\midrule
Precision & 0.9326 \\
Sensitivity (Recall) & 0.9326 \\
Specificity & 0.9372 \\
Accuracy & 0.9350 \\
\bottomrule
\end{tabular}
\caption*{Performance Metrics}
\label{tab:performance_metrics}
\end{table}

\begin{figure}[H]
    \includegraphics[width=0.5\textwidth]{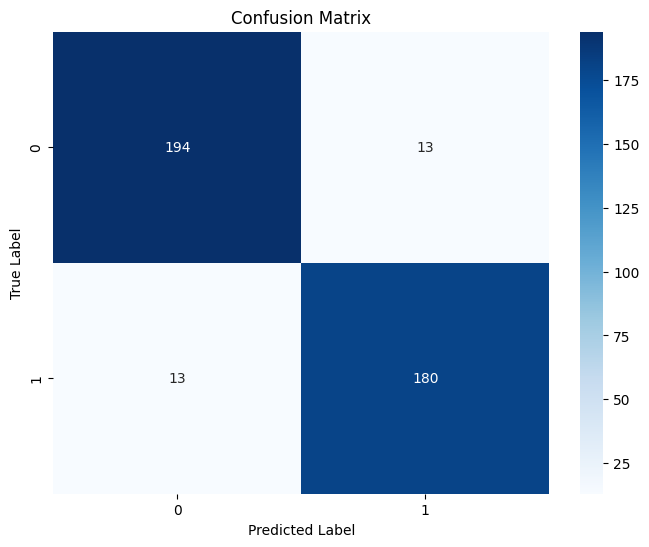}
    \caption{Confusion Matrix}
    \label{Flowchart}
\end{figure}

\subsection{Experimental Setup and Database Collection}
The dataset comprises of 2000 videos of Violence and NonViolence each. The videos were collected and compiled from CCTV footages, violence recordings available on YouTube and made available on Kaggle.
The nonviolent videos also cover the cases of Human interactions like handshakes and hugging which shall not fall under the category of Violence.

The dataset , comprising of 4 GB of data divided into 2 classes of 2 GB each and 1000 videos in each. The two classes were violence and Non Violence. Of the entire dataset, 72 percent was used for Training, 8 percent as Validation Dataset and 20 percent as testing dataset, the elements in the three sets of data did not overlap, implying that testing data was totally different from training data and the validation data.
\subsection{Table, Graph with description}
The following graphs give an idea about have the training and validation data witnessed varying accuracy and loss with subsequent epochs. Validation and Training Accuracy were found to increase with the increasing number of epochs while loss was found to be decreasing for both and subsequently reached 0.226 for validation. These imply that the model is learning and it is improving with more cycles of training.
\begin{figure}[H]
    \includegraphics[width=0.5\textwidth]{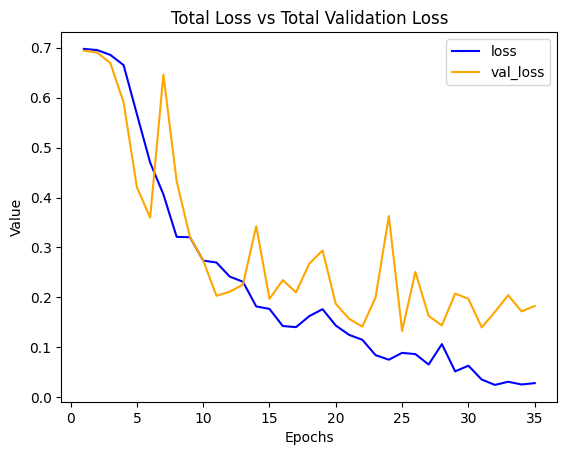}
    \caption{Loss}
    \label{Flowchart}
\end{figure}

\begin{figure}[H]
    \includegraphics[width=0.5\textwidth]{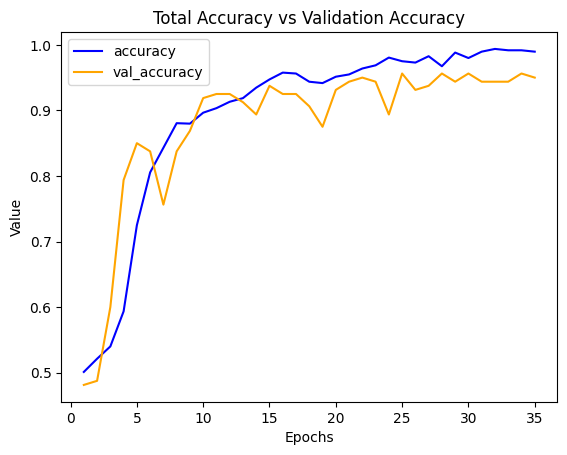}
    \caption{Accuracy}
    \label{Flowchart}
\end{figure}

\subsection{Result comparison}
In view of two similar research works already existing with the 3D CNN and the 2D CNN with Hough Forests (The references to these works are attached in references), their documented Accuracies in comparision to our model's accuracies.
\begin{table}[H]
\centering
\begin{tabular}{lc}
\toprule
\textbf{Model} & \textbf{Accuracy} \\
\midrule
3D-CNN & 65\% \\
2D-CNN + Hough forests & 92\% \\
Proposed Solution with BiLSTM & 93.5\% \\
\bottomrule
\end{tabular}
\caption*{Model Accuracies}
\label{tab:model_accuracies}
\end{table}

\section{Conclusion and Future works}
In this paper we study how effective is 2D-CNN and BiLSTM based Fight Scene Detection Model. The model was trained and tested on the Violence dataset. the performance was evaluated using unseen part of the dataset along with the cross validation procedures. From the results , it is observed to have an accuracy of 93.5 \% for unseen dataset. Thereafter the model generates highlight compiling the predicted violent scenes from the input video. Future works include improving system specifications and optimising the code to reduce time complexity and giving it the ability to feed longer videos and operate more efficiently. In future works, we also focus on using pretrained weights of imagenet  and increasing the training dataset variety to improve the prediction in varied cases.
\section{Relevant}


\begin{thebibliography}{00}
\bibitem{}
You can also download and access it on drive from here: The dataset is named as archive.\\
bit.ly/3w911z1
\end{thebibliography}
\end{document}